# Large language models improve physician accuracy but lead to false reliance

Tirtha Chanda[1,2], Christoph Wies[1,2], Franziska Schramm[1], Carina Nogueira Garcia[1], Nicolas B. Merl[1,2], Martin J. Hetz[1,2], Jochen S. Utikal[3,4,5], Phillip Tschandl[6], Cristian Navarrete-Dechent[7], Alexander Thiem[8], Jakob N. Kather[9], Consortium*, Titus J. Brinker[1*]

1. Division of Digital Prevention, Diagnostics and Therapy Guidance, German Cancer Research Center (DKFZ), Heidelberg, Germany
2. Medical Faculty, University Heidelberg, Heidelberg, Germany
3. Skin Cancer Unit, German Cancer Research Center (DKFZ), Heidelberg, Germany
4. Department of Dermatology, Venereology and Allergology, University Medical Center Mannheim, Ruprecht-Karl University of Heidelberg, Mannheim, Germany
5. DKFZ Hector Cancer Institute at the University Medical Center Mannheim, Mannheim, Germany
6. Department of Dermatology, Medical University of Vienna, Vienna, Austria
7. Department of Dermatology, Escuela de Medicina, Pontificia Universidad Católica de Chile, Santiago, Chile
8. Clinic and Polyclinic for Dermatology, Venereology and Allergology, University Medical Center Rostock, Rostock, Germany
9. Department of Medical Oncology, National Center for Tumor Diseases (NCT), Heidelberg University Hospital, Heidelberg, Germany

*Corresponding author:
Titus J. Brinker: titus.brinker@dkfz.de

## Abstract

Retrieval-augmented large language models (LLMs) promise source-linked clinical support, but their value depends on whether displayed evidence guides rather than distorts physician reliance. We developed CORA, an agentic retrieval-augmented LLM, to investigate how source-linked assistance affects physician decision-making. CORA maintained benchmark performance and achieved larger gains on cases published after the models' training-data cutoffs. In a study of 46 physicians, accuracy increased from 70.8% unaided to 82.6% with CORA. Supporting citations predicted correct answers (87.7% vs 65.5%), but citations created an important asymmetry: perceived support increased adoption of correct advice from 34% to 76.9% but when an incorrect LLM answer appeared citation-supported, physician resistance to it fell from 92% to 34.8%. These findings show that source-linked LLM assistance can improve physician accuracy while introducing a grounding-dependent safety risk.

# Main

Large language models (LLMs) have shown strong performance on medical tasks, processing complex clinical narratives and generating clinically relevant responses[1–3]. Yet they remain prone to generating fluent but factually inaccurate outputs, may fail to reflect evolving clinical guidelines, and offer limited transparency regarding the evidentiary basis of their responses[4,5]. Retrieval-augmented generation (RAG) mitigates these limitations by supplying an LLM with external documents at inference time[6,7]. Agentic RAG extends it by iteratively evaluating whether retrieved evidence is sufficient, reformulating the query, and repeating retrieval when needed[8–10]. However, retrieval introduces a failure mode distinct from inaccuracy: the generated answer may not reflect the retrieved sources, even when it carries apparently relevant citations[11].

Evaluations of LLM-assisted clinical reasoning have reported mixed effects on physician performance[12–14]. Meanwhile, studies of medical RAG systems have measured the accuracy of standalone model outputs or clinicians' ratings of those outputs[15–19]. These two research areas have developed largely in parallel. One focuses on the efficacy of LLM support in enhancing clinical decisions, while the other examines the accuracy and quality of retrieval-based outputs. Neither line of work investigates what happens when a physician actually reads and judges support of the cited evidence and uses it to make a decision. Whether retrieval-based LLM assistance improves physician performance, and whether perceived citation support promotes appropriate reliance or harmful deference to incorrect advice, therefore remains untested.

Here we developed CORA (Citation-Oriented Retrieval Assistant), an agentic RAG system (Fig. 1a), to investigate whether retrieval-based LLMs improve physician question answering performance and how its cited evidence influences physician reliance. We first compared CORA with non-retrieval baselines across multiple LLMs (such as GPT-5, Llama-4, etc.) on a dataset compiled from four dermatology question-answering benchmarks, testing if retrieval improved model performance. We then repeated the evaluation on a dataset built from dermatology case reports published after each model's training cutoff, testing if retrieval benefits held on cases that could not have been seen during model training. Next, we conducted a reader study in which 46 physicians answered dermatologic questions unaided, then viewed CORA's answer and its citations, rated how well the citations supported that answer, and could revise their response (Fig. 1b). This allowed us to investigate whether CORA-assistance improved physician accuracy, whether CORA's cited sources supported its own answers, and how perceived citation support related to physician answer revision and their reliance on correct versus incorrect advice. Together, these analyses connect model accuracy, evidentiary support, and physician decision-making, revealing that the same signal of citation support that promotes appropriate reliance can also drive deference to incorrect LLM advice when the cited evidence appears to support it.

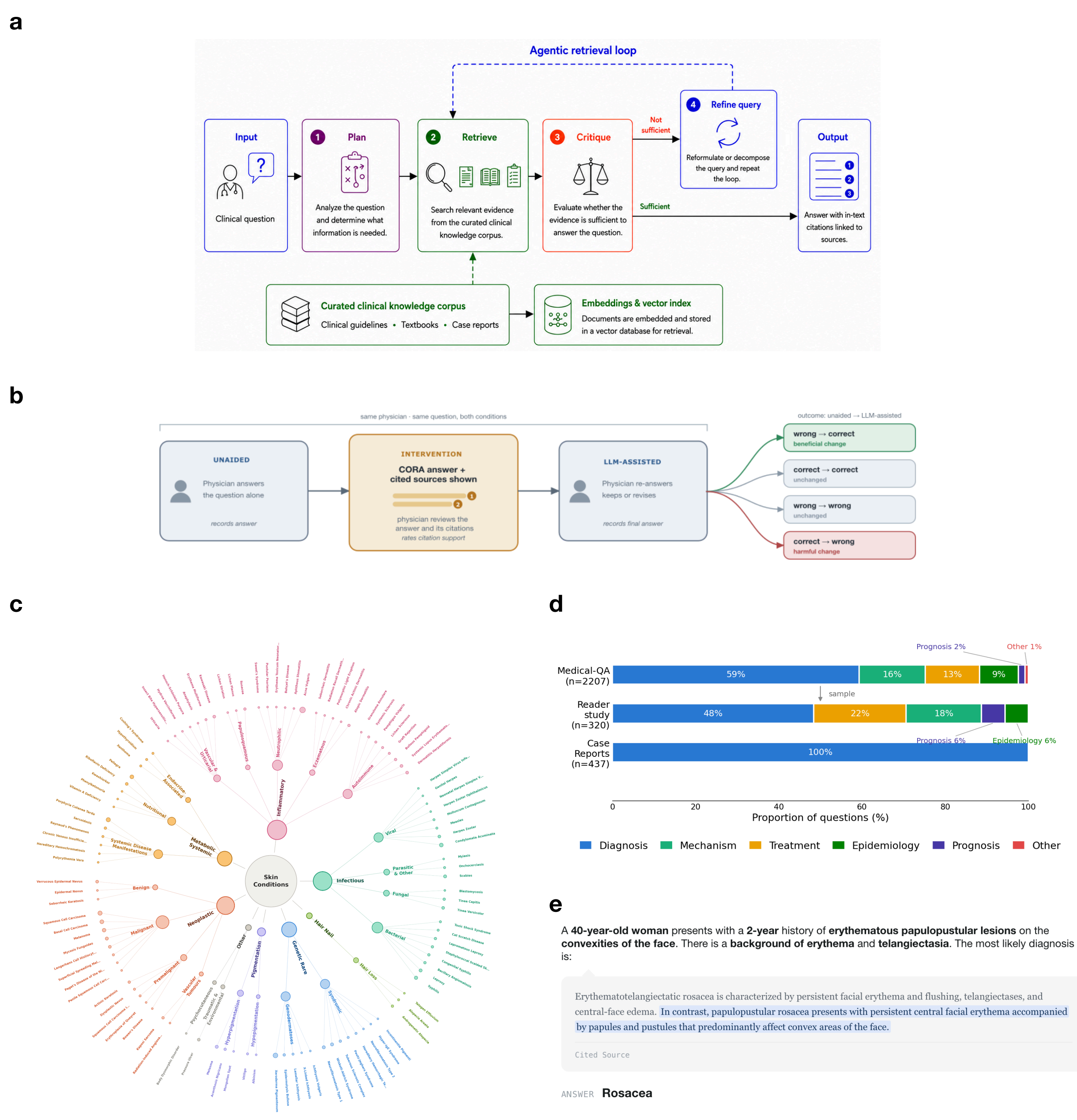

## Figure 1: The CORA citation-grounded retrieval assistant and reader-study design.

**a**, CORA architecture and agentic retrieval loop. For each clinical question, CORA plans and retrieves evidence from a vector database of embedded clinical guidelines, textbooks, and case reports. It evaluates evidence sufficiency and, if needed, refines the query based on knowledge gaps in the retrieved evidence, and repeats retrieval before generating an answer with linked citations. **b**, Within-subjects reader-study schema. Each physician answers a given question under two conditions. Unaided, the physician records an answer alone. In the intervention step, CORA's answer and its cited sources are displayed; the physician reviews the answer and citations and rates per-source citation support for the CORA answer. The physician records a final answer, keeping or revising the initial one. **c**, Coverage of the dermatology question sets. Circular dendrogram of skin conditions organized into higher-order categories with node size scaled to the number of questions per condition. **d**, Distribution of clinical question types across the three evaluation sets: the DermBenchQA dataset (multiple-choice, n=2,207), the stratified reader-study sample (n = 320) and the contamination-resistant DermCaseQA set (open-ended, n=437). Bars show the proportion of items in each category. **e**, Representative reader-study item. A clinical vignette (here, erythematotelangiectatic versus papulopustular rosacea) is shown alongside the source CORA cites in support of its answer, illustrating the citation-grounded format physicians evaluated.

# Results

We evaluated CORA on two compiled dermatology question sets. The first, DermBenchQA, comprised 4,855 single-best-answer questions drawn from four medical question-answering (QA) benchmarks. The second, DermCaseQA, comprised 998 open-ended questions generated from 893 PubMed-indexed dermatology case reports published in 2025 or later, after the training cutoff of every backbone model evaluated. The two sets together spanned 591 conditions across eight disease categories which represents the breadth of dermatological practice from inflammatory and infectious disease to neoplastic, genetic, and pigmentary disorders (Fig. 1c).

We conducted a within-subjects reader study where 46 physicians answered a 320-item subsample of DermBenchQA, 16 questions per physician, with CORA's answers generated by an open-weight model, Llama-4 Scout. The dataset's five clinical question types were all represented, with diagnosis and treatment questions oversampled (Fig. 1d). Each physician first recorded an answer unaided. They were then shown CORA's answer together with its retrieved citations (Fig. 1e) and asked to judge whether the cited sources supported or did not support the CORA's answer. They were instructed to assess citation support independently of whether they believed CORA's answer to be correct, and this instruction was restated above every rating item to prevent drift over the course of the session. They could then record a final answer, either retaining or revising their initial choice. Each question therefore contributes a paired unassisted/assisted decision. Details on the sampling procedure and CORA configurations are provided in Methods.

## CORA performance across LLMs

We evaluated CORA against each base model on DermBenchQA, restricting the comparison to questions where the retriever reached sufficiency within three iterations (2,207 of 4,855; 45.5%) so that both systems answered an identical set of items. We framed the primary comparison as a non-inferiority test since CORA adds traceable source citations, a function absent from the base models; preserving baseline accuracy while adding traceability would therefore constitute a meaningful advantage. Across all five backbone LLMs, CORA maintained performance to the corresponding baseline on the 2,207 matched DermBenchQA items (one-sided Wald non-inferiority test, margin 1 percentage point (pp); all Holm–Bonferroni corrected P-values ≤ 0.004; Fig. 2a). Accuracy held or improved in every case, but the magnitude of improvement scaled inversely with baseline capability: GPT-5, already near ceiling, gained 0.3 pp (92.7% to 93.0%; 95% CI: -0.7, 1.2), whereas the weakest backbone, Gemma 3, gained 8.6 pp (76.9% to 85.5%; 95% CI: 6.9, 10.2), with intermediate models in between: Qwen 2.5 3.9 pp (95% CI: 2.5, 5.2), Mistral Large 2 1.8 pp (95% CI: 0.3, 3.4) and Llama 4 1.4 pp (95% CI 0.2, 2.7) (Fig. 2a).

Category-resolved deltas mirrored this pattern. Gains were higher in lower-baseline, lower-prevalence categories, such as metabolic/systemic (up to 14 pp gain for Qwen 2.5), genetic/rare (12 pp gain for Gemma 3) and infectious disease (11 pp gain for Gemma 3), while high-baseline common categories such as inflammatory dermatoses moved little or occasionally declined (e.g., 1 pp loss for GPT-5). The few negative deltas confined to rare categories (e.g., hair/nail, n = 38) were not consistent across all models (Fig. 2b).

To test whether these gains reflected reasoning over retrieved text rather than exposure to public dataset items during model pretraining, we repeated the comparison on the training data contamination-resistant DermCaseQA set, again restricting to sufficiency-reached items (437 of 998; 43.8%). CORA was again non-inferior to baseline across all models (one-sided Wald non-inferiority test, margin 1 pp; all P ≤ 0.005, Holm–Bonferroni corrected; Fig. 2c).

Within this set, retrieval gains were larger and more uniform than on DermBenchQA. Increase again scaled inversely with baseline capability: Gemma 3 improved by 22.7 pp (37.5% to 60.2%; 95% CI: 17.7, 27.6), Qwen 2.5 by 18 pp (46.5% to 64.5%; 95% CI: 13.1, 23.0), Llama 4 by 10.9 pp (51.3% to 62.2%; 95% CI: 6.8, 15.2) and Mistral Large 2 by 9.9 pp (55.1% to 65.0%; 95% CI: 5.2, 14.4), while the strongest backbone, GPT-5, still improved by 3.4 pp (77.1% to 80.5%; 95% CI 0.1, 6.8; P=0.005) (Fig. 2c). Category-resolved gains were positive in nearly every cell (Fig. 2d), with the largest uplift in neoplastic (up to 22 pp gain), genetic/rare (up to 26 pp gain) and metabolic/systemic (up to 33 pp gain) presentations. The isolated negative deltas occurred only in the smallest strata (metabolic/systemic, n = 21; other, n = 23) and were not reproduced across models. Because the two sets differ in answer format (single-best-answer versus free text), accuracy is not directly comparable across them, and we treat them as separate axes of evaluation rather than a pooled score.

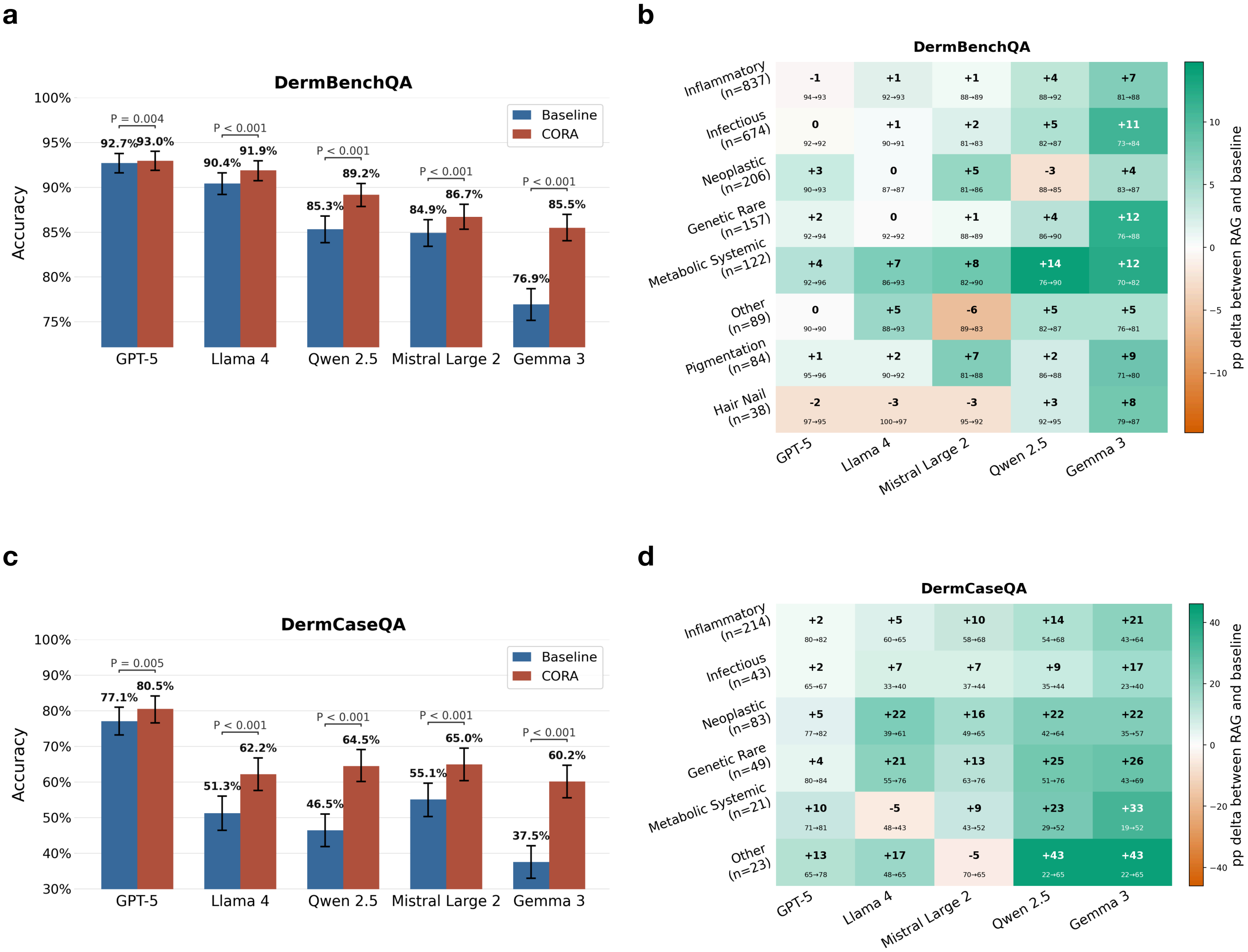

**Figure 2: Agentic retrieval improves LLM accuracy across models and dermatology subdomains.**

**a**, DermBenchQA dermatology dataset accuracy for five base models (GPT-5, Llama 4, Qwen 2.5, Mistral Large 2 and Gemma 3), comparing each model's baseline (no retrieval) with CORA (retrieval-enabled). Bars show accuracy on n = 2,207 matched multiple-choice questions per model; error bars denote 95% confidence intervals. P values are from one-sided Wald non-inferiority tests (margin=1 pp), Holm–Bonferroni corrected across models (m=5). **b**, Disease category effect of CORA on DermBenchQA, shown as the percentage-point (pp) change in accuracy relative to baseline for each model × condition category. Rows are disease categories (with question counts); the large number and cell colour give the CORA minus baseline difference in pp, and the small numbers give baseline accuracy to CORA accuracy on the paired questions. **c**, Training data contamination-resistant case reports dataset DermCaseQA (open-ended, n=437 matched questions per model), plotted as in a. Retrieval gains are substantially larger than on the DermBenchQA set. **d**, Disease category effect of CORA on DermCaseQA, plotted as in b. Note the wider colour scale, reflecting the larger effect sizes on this dataset.

## CORA assistance improves physician accuracy

Participants were 46 physicians in dermatology from 21 countries, including 3 dermatology residents. Among the 43 who had completed residency, post-residency experience was <1 year (n=2), 1-3 years (n=8), 4-9 years (n=13), and ≥10 years (n=20). Across 46 physicians and 736 physician-question decisions, mean accuracy increased from 70.8% (95% CI 67.0%, 74.5%) without assistance to 82.6% (95% CI 80.0%, 85.2%) after participants viewed CORA's answer and cited sources (Wilcoxon signed-rank test, P<0.001, n=46 physicians; Fig. 3a). 37 physicians improved, 8 were unchanged, and 1 declined. The direction of effect was consistent across all question types and disease categories (Fig. 3b). At the decision level, assistance corrected 110 initially incorrect responses and overturned 23 initially correct responses into incorrect; 498 responses remained correct and 105 remained incorrect (Fig. 3c). The aggregate benefit was accompanied by a smaller but clinically relevant number of harmful reversals. Physician performance by experience levels are provided in Extended Data Fig. 1.

## Citation support is associated with model accuracy and physician error correction

To derive a question-level measure of citation support, we aggregated physicians' ratings by majority vote: a question was classified as supported when 2 out of 3 physicians who evaluated it judged at least one displayed citation to support CORA's answer (inter-rater agreement in Extended Data Fig. 2). Across the 192 questions rated by three physicians, at least one citation was judged supportive for 80.2% (citation support rate; 95% bootstrap CI: 74.5%, 85.4%), while 60.6% of individual cited sources were rated as supportive (citation precision; 95% CI: 56.2%, 64.9%).

The presence of at least one citation supporting the generated answer was strongly associated with CORA's correctness (Fig. 3d). Using the majority-voted question-level citation support, CORA answered 92.9% of questions correctly when at least one citation supported its answer, compared with 65.8% when no citation was supportive (odds ratio (OR), 6.76; 95% CI: 2.73, 16.77; P<0.001; n=192 questions).

This association extended to physicians' final decisions, analysed at the level of each individual physician's own citation ratings rather than by majority vote. Final answers were correct in 87.7% of responses when at least one citation was judged to support the CORA answer, compared with 65.5% when no citation was judged supportive (OR, 3.75; 95% CI: 2.48, 5.67; P<0.001, n=736 physician-question decisions).

The relationship was particularly pronounced when physicians' initial answers were incorrect, identifying the pathway underlying the beneficial changes. Physicians recovered from initially incorrect answers specifically when they judged CORA's citations to support its answer. In these cases, physicians reached the correct final answer in 64.6% of responses when at least one citation supported the CORA answer, compared with 23.9% when no supporting citation was present (OR, 5.79; 95% CI: 2.82, 11.90; P<0.001; n=215 physician-question decisions). Thus, citation support identified not only more accurate CORA outputs but also circumstances in which physicians were significantly more likely to correct an initially incorrect decision.

## Perceived citation support improves adoption of correct advice but weakens resistance to incorrect advice

Although CORA improved physician accuracy overall, aggregate performance conflates two behaviours with opposite safety implications: appropriately adopting correct CORA advice and appropriately resisting incorrect advice. An intervention can increase mean accuracy while worsening the more dangerous error class, physicians abandoning a correct answer for an incorrect one. To characterise how physicians used CORA, we applied the appropriate-reliance framework of Schemmer et al[20]. which distinguishes two beneficial behaviours: accepting AI advice when it is correct, and rejecting AI advice when it is not. Among decisions where physicians were initially incorrect and CORA was correct, relative AI reliance (RAIR) is defined as the proportion they revised to CORA's answer. Among decisions where physicians were initially correct but CORA was incorrect, relative self-reliance (RSR) is defined as the proportion who retained their own correct answer. Restricting each rate to cases where reliance could change the outcome separates reliance behaviour from the base rate of who was correct to begin with. Because both behaviours are desirable, well-calibrated reliance would drive both rates toward 100%. Physicians adopted correct CORA advice in 64.3% of cases where they were wrong (RAIR; 95% CI: 56.9%, 71.4%) and resisted incorrect CORA advice in 64.6% of cases where they were right (RSR; 95% CI: 51.0%, 78.4%), placing them in the appropriate-reliance quadrant (Fig. 3e). Reliance was therefore discriminating rather than indiscriminate deference, the pattern described as automation bias would produce high RAIR with RSR approaching zero. However, this reliance was not calibrated, and neither aggregate rate reveals when self-reliance failed.

Stratifying these behaviours by perceived citation support revealed a directional dissociation between the two forms of reliance (Fig. 3f). When CORA was correct, RAIR was higher among decisions in which at least one citation was judged supportive (93/121; 76.9%; 95% CI: 67.8%, 84.9%) than among those in which no citation was judged supportive (17/50; 34.0%; 95% CI: 20.8%, 47.7%; 42.9 pp increase). When CORA was incorrect, the association reversed: RSR was lower with perceived citation support (8/23; 34.8%; 95% CI: 17.2%, 52.6%) than without it (23/25; 92.0%; 95% CI: 79.2%, 100%; 57.2 pp decrease). Concretely, when physicians judged that a citation supported an incorrect CORA answer, they abandoned an initially correct response in 15 of 23 cases. Thus, the same perceived grounding signal that promotes beneficial reliance may also create a key safety risk: a grounding-miscalibration failure mode that arises when physicians read a citation as supporting an incorrect answer. Examples of incorrect CORA outputs that appear supportive are provided in Supplementary Note 1.

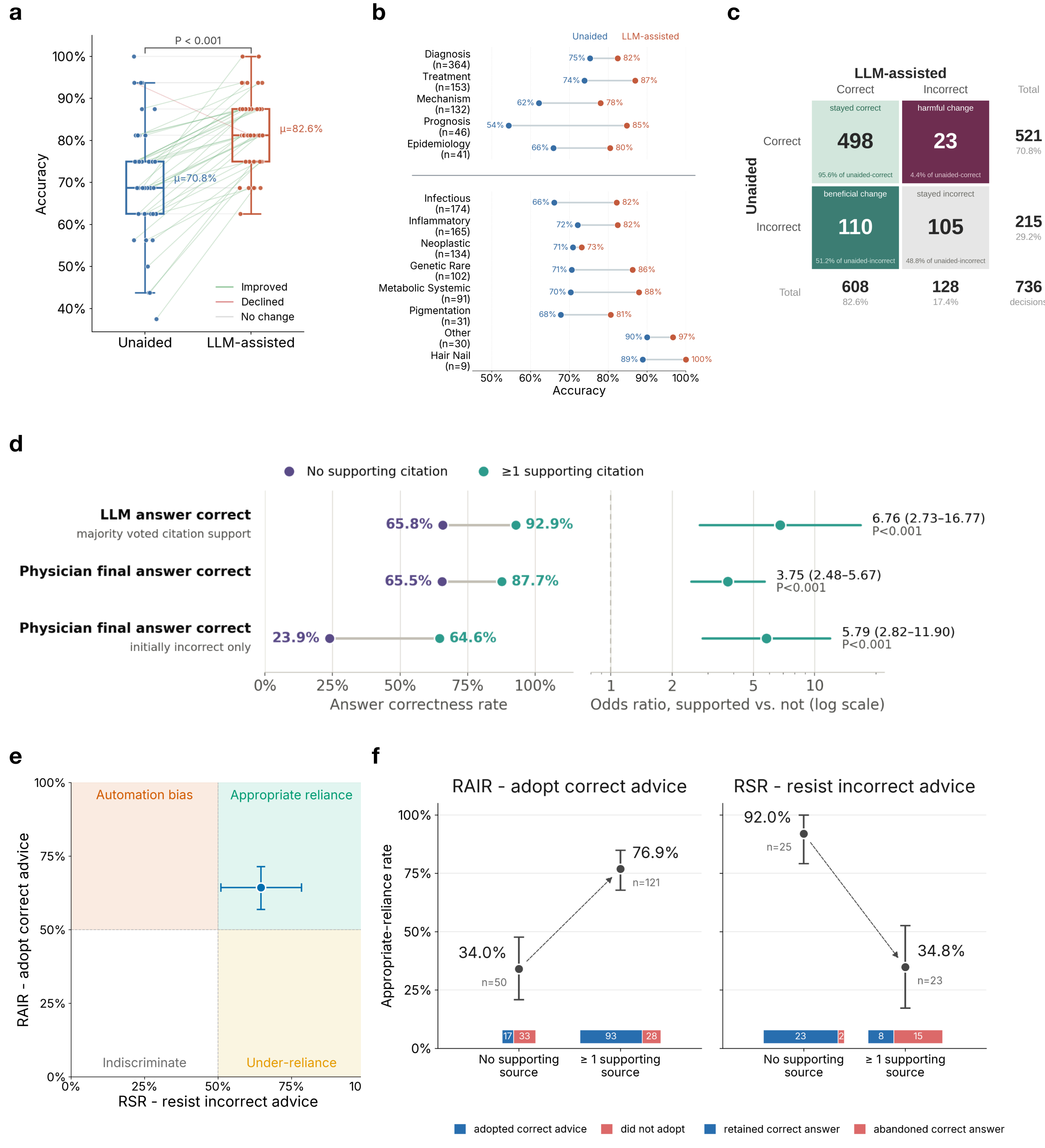

**Figure 3: CORA assistance improves physician diagnostic accuracy, but the appropriateness of reliance is governed by perceived citation support.**

**a**, Per-physician decision accuracy in the unaided versus LLM-assisted condition (n=46 physicians, 736 decisions). Each connected pair of points represents one physician; line colour denotes whether assisted accuracy improved, declined or was unchanged relative to the unaided baseline. Box plots show the median and interquartile range, with whiskers extending to 1.5× IQR. Pooled accuracy increased from 70.8% (95% CI 67.0%, 74.5%) unaided to 82.6% (95% CI 80.0%, 85.2%) with LLM assistance (Wilcoxon signed-rank test, $P<0.001$, n=46 physicians). **b**, Unaided versus LLM-assisted accuracy within prespecified subgroups defined by question type and disease category with point estimates. **c**, Paired 2×2 table of 736 case-responses by correctness when answered unaided (rows) versus with LLM assistance (columns). In-cell percentages are row-normalised; margins are shares of all decisions. Beneficial changes outnumbered harmful ones nearly five-to-one. **d**, Physician-judged citation support versus correctness. Participants rated each source cited with the LLM answer as "supports" or "does not support" it; an answer is supported if ≥1 cited source was rated as supporting. Left: correctness rates for unsupported versus supported answers, for three outcomes: LLM answer correct, physician final answer correct, and physician final answer correct among decisions where their unaided answer was incorrect. Right: corresponding odds ratios (supported vs unsupported) with 95% CIs on a log scale; all $P<0.001$. The LLM-correctness row is question-level (192 questions rated by three physicians, support derived by majority vote); the two physician rows are response-level (n=736 and n=215) with logistic regression clustered by physician. **e**, Appropriateness-of-reliance map (Schemmer AoR framework). Relative AI reliance (RAIR), adoption of correct advice when the physician was incorrect, was 64.3% (95% CI: 56.9, 71.4). Relative self-reliance (RSR), resistance to incorrect advice when the physician was correct, was 64.6% (95% CI: 51.0, 78.4). The single point marks the pooled estimate across all questions where physicians and the LLM answer differed. Whiskers are 95% cluster-bootstrap CIs (10,000 resamples of physicians). Quadrants partition the space into appropriate reliance, over-reliance and under-reliance. **f**, The grounding-miscalibration effect, shown for RAIR (adopt correct advice) and RSR (resist incorrect advice), each stratified by citation support. Points show the stratum estimate with 95% CI; per-stratum denominators are indicated below each estimate. When LLM advice carries a supporting citation, physicians become more likely to adopt correct recommendations (RAIR increases from 34% to 77%) but less likely to resist incorrect ones (RSR decreases from 92% to 35%).

# Discussion

In our study involving 46 physicians and 736 decisions, agentic retrieval-augmentation increased physician accuracy 11.8 percentage points, and physicians adopted correct advice on 64.3% of the cases they had initially answered incorrectly. But the more informative finding was when physicians changed their minds. Whether a physician judged that at least one displayed citation supported CORA's answer was associated with almost everything else we measured. Physicians were significantly more likely to reach the correct final answer on questions for which they judged CORA's citations to support its response, particularly on cases they had initially answered incorrectly. But this favourable average concealed a stratified failure. The same perceived support that accompanied physicians overturning their own errors also accompanied their abandonment of correct answers when CORA was wrong. Together, these findings show that the clinical value and safety of retrieval-based LLM assistance depend not only on whether the model is correct, but also on whether the evidence presented to clinicians genuinely supports its recommendation.

In the benchmark evaluation, CORA was noninferior to each unaugmented backbone on cases meeting the prespecified retrieval-sufficiency criterion, with larger gains for smaller models and for dermatology case reports published after every evaluated model's training cut-off. Such cases are also among those most likely to prompt a physician to consult a decision-support system in the first place.

Post-assistance accuracy is therefore an incomplete measure of clinical utility. Appropriate reliance requires adopting advice when it is correct and resisting it when it is not, and aggregate improvement can hide newly introduced errors. Stratifying reliance by perceived citation support separated these two behaviours. We term the resulting pattern grounding miscalibration: perceived evidentiary support that is not reliably aligned with the correctness of the answer it accompanies. The failure is not in physicians' judgement of citations but in the fact that the display of citations made an incorrect answer look grounded. A citation that is topically relevant but does not establish the generated claim may thus be more dangerous than an unsupported answer, because it manufactures an appearance of verifiability that can lead a physician to abandon a correct assessment.

Our findings connect multiple largely separate literatures. Work on clinical automation bias has long shown that erroneous automated advice can lead clinicians to overturn correct decisions[21,22], including a study in which flawed LLM suggestions reduced accuracy despite AI-literacy training[23]. In those studies, LLM advice was not accompanied by displayed evidence, leaving open whether deference is a general response to automated recommendations or is conditioned on what is shown alongside them. We find it is conditioned: resistance to incorrect recommendations fell according to whether physicians judged the accompanying citations supportive. Studies of human-AI interaction indicate that explanations can increase reliance without improving joint decision quality, amplifying deference when the recommendation is wrong[24,25]. Heatmap-based imaging AI explanations accelerated radiologists' deference to both correct and incorrect diagnoses[26]. Citations increased self-reported trust in LLM-generated answers even when the citations were randomly selected[27]. Those citation effects were stated trust in non-expert samples, whereas we measured physicians' revisions of decisions based on citations accompanying LLM answers. Physicians in the assisted condition did not exceed CORA's own accuracy. This reproduces in a citation-based LLM a pattern previously reported for LLMs without citations[12,13], which our reliance stratification analysis may explain. Output-level assessments, meanwhile, show that model-generated medical citations frequently fail to support, or even contradict, the claims they accompany[11], without investigating what such failures do to clinical decisions. Our work supplies the missing downstream link: the unreliable citations identified in such assessments are the ones most likely to reduce physician scrutiny, because their presence is treated as a proxy for correctness. Two safeguards fail at once: LLMs lack the metacognition to flag when their outputs should be distrusted[28], and the citations attached to those outputs carry no compensating signal. The signal physicians do have is not absent but asymmetric: it served them well when they judged the evidence unsupportive, and failed them when they judged it supportive.

Two implications follow for how these systems should be developed and evaluated. First, accuracy alone is an inadequate endpoint. A system can raise average accuracy while making its residual errors harder to catch, and only reliance-stratified reporting will reveal this. A systematic review of 519 studies on LLMs in healthcare found that over 95% used accuracy as the primary evaluation[29]. Evaluations of retrieval-augmented clinical tools should report performance separately for cases in which clinician and system disagree, and specifically for the subset in which the system is wrong. Second, the design target is the gap between how supportive a citation appears and how much it actually establishes the answer. A source can be on-topic (right disease, right drug) without containing the specific evidence that makes the answer correct, and physicians in our study were not reliably telling these apart. Designs worth testing therefore include systems that flag when retrieval was weak, that indicate whether a source proves the answer rather than simply matching its subject, and that require an explicit check before an answer is revised.

Neither implication is likely to be specific to citations or to dermatology. Saliency overlays, retrieved sources, and confidence scores all function as evidence cues, and the radiology findings noted above suggest that cue presence can govern reliance independently of cue validity. Retrieval-grounding is widely treated as the principal safeguard against LLM error in clinical settings, on the assumption that a displayed source invites verification[30]. Our results indicate that this assumption cannot be taken for granted: the cue that is meant to enable scrutiny can instead substitute for it.

Several limitations qualify these conclusions. The absence of citation-free and non-retrieval conditions prevents isolating the components responsible for the accuracy gain, and the sequential design cannot separate CORA’s effect from generic reconsideration. Model comparisons were restricted to questions meeting the retrieval-sufficiency criterion, so the results reflect CORA's performance when the corpus contains relevant evidence, not on arbitrary queries. The post-cutoff evaluation set, while contamination-resistant, comprises published case reports, which are selected for instructiveness and may not represent routine presentations. The findings rest on responses to structured vignettes rather than live clinical encounters, which may not fully reflect how physicians weigh LLM recommendations under real workflow conditions, and participants were instructed not to use other AI tools, which might have contradicted the displayed evidence. The self-reliance findings rest on small strata: the resistance analysis comprised 48 decisions, of which 23 involved citation-supported recommendations. Because incorrect recommendations were uncommon and citation support was rated rather than experimentally manipulated, the harmful-reliance estimates are imprecise and observational.

Separating these effects will require experiments that vary answer correctness and citation quality separately. Correct and incorrect recommendations could be paired with evidence that genuinely supports, simply matches the topic, or contradicts it, as well as with no citations. This would show whether apparent support directly increases reliance and which part of the pipeline produced the observed benefit. For now, our findings support a narrower conclusion: physicians' judgement that a citation supported the answer predicted whether they deferred to it. This held even when CORA was incorrect, in which case no cited source could have truly supported it. Retrieval-based clinical LLMs should therefore be evaluated not only on answer accuracy and retrieval quality, but on how their presentation of evidence shapes reliance on correct and incorrect advice, and designed so that the support a citation appears to provide tracks the support it actually provides.

# Methods

## Ethics statement

The study's ethics vote was held by the University Clinic Mannheim of the Medical Faculty of the University of Heidelberg, and was conducted in accordance with the Declaration of Helsinki. All participants provided written informed consent through a consent form embedded at the start of the study describing study purpose, voluntary participation, the right to withdraw, data handling. No patient data were involved at any stage; all data consisted of published multiple-choice questions and published case reports. Physician responses were pseudonymised and the linkage key was held by the study coordinators in a password-protected file. All analyses were performed on pseudonymised data.

## Design overview

We developed CORA (Citation-Oriented Retrieval Assistant), an agentic retrieval-augmented generation (RAG) system for dermatological decision support that returns an answer together with the source texts on which that answer is based, and evaluated it in two stages. In the first stage, we benchmarked CORA against unaugmented baselines across five large language models (LLMs) on a dermatology multiple-choice question set and on a contamination-resistant open-ended question set derived from case reports published after the training cut-off of every evaluated model. In the second stage, we conducted a preregistered within-subjects reader study in which 46 physicians answered dermatology multiple-choice questions unaided and then with CORA assistance, rating the support of each cited passage before being permitted to revise their answer. The benchmark establishes whether retrieval augmentation preserves or improves answer accuracy; the reader study establishes how physicians calibrate their reliance on CORA's output, and constitutes the primary contribution of this work. Throughout, "CORA" denotes the full pipeline: knowledge base, agentic retriever, reranker, and generation model, and "baseline" denotes the same generation model queried without retrieved context.

## Datasets

### DermBenchQA

We compiled a dermatology-focused multiple-choice question dataset of 4,855 questions from four publicly available medical benchmark sources: MedMCQA[31] (n=3,412), MedQA[32] (n=1,411), MMLU[33] (n=19), together with 13 items from the written state examination administered by the Institut für medizinische und pharmazeutische Prüfungsfragen (IMPP). The MedQA total includes 580 stems distributed in both a five-option and a reduced four-option format; because distractor count affects item difficulty, both formats were retained as distinct items. Questions were filtered for relevance to dermatology using an LLM with the prompt in Supplementary Table 1 followed by manual review of a sample of 30 questions, of which none were removed. Questions requiring visual reasoning (e.g., referencing an image not present in the dataset) were retained but flagged.

### DermCaseQA

To complement the multiple-choice dataset with a training data contamination-resistant format, we built an open-ended set of 998 questions from 893 dermatology case reports whose text first became public on or after 1 January 2025, after the training cut-off of every evaluated model. Candidates were retrieved from the PubMed Central open-access subset (Case Reports[pt], Skin Diseases[mesh], Humans[mesh], year ≥ 2025; n = 1,372).

Because the recorded year is usually the print issue date, a 2025-stamped report may have been readable much earlier as an ahead-of-print article or a preprint, and that earlier text is what an LLM could have been trained on. We therefore judged each of the 1,372 candidate reports by its earliest date on which the text was demonstrably public, rather than by its recorded year. Using the PubMed and Crossref records for every report, we took the earliest of all dates indicating public release: electronic ahead-of-print and PubMed listing dates, Crossref online and print publication dates, and the posting date of any corresponding preprint. Dates reflecting private editorial steps, manuscript submission, acceptance, and revision, and Crossref's internal deposit timestamp, were disregarded, since the text was not yet public at those points. Preprints were identified both from explicit publisher-deposited links between an article and its preprint and by searching Crossref for preprint records with a near-identical title. A report was excluded if its earliest public date, or any associated preprint, predated the 1 January 2025 cutoff. This removed 96 of 1,372 reports (7.0%).

Each retained report was reformulated into a vignette using Claude Sonnet 4.6, preserving the presenting features while removing the published diagnosis, which served as the reference answer (n = 1,538 candidates). A verifier LLM (Claude Sonnet 4.6), given the source text but blind to the key, answered each item itself and screened it for a single defensible answer, answer leakage, groundedness in the source, and answerability without a figure; items passing all four checks, matched by the verifier's own answer, and not duplicating a sibling item's diagnosis were retained (477 removed, 457 of them for leakage). Dropping 63 further items whose source failed the date gate left 998 items from 893 reports.

## Question categorisation

Every question in both datasets was annotated by Claude Sonnet 4.6 to support stratified sampling and subgroup analysis. Four independent passes recorded (i) clinical content - named condition, disease category, estimated prevalence in routine dermatological practice, question type, and whether visual reasoning was required; (ii) stem structure, classified as a one-liner, short stem, or vignette; (iii) question language and unambiguous structural defects such as duplicated or missing options, with the keyed answer withheld and calibration set to flag only clear defects; and (iv) difficulty (easy, medium, hard) from the perspective of a competent board-certified general dermatologist, with the keyed answer shown. Free-text disease names were normalised to canonical labels by hierarchical deduplication and each label assigned one of 29 predefined medical subcategories, constraining the model to a closed label set. Questions were annotated independently of one another and the same prompts were applied to both datasets. Prompts, category definitions, and the normalisation procedure are given in Supplementary Notes 2.1–2.6; full distributions in Supplementary Tables 2, 3, and 4.

## CORA development

### Knowledge base construction

The domain knowledge base was indexed in a vector database ChromaDB (version 0.6) and comprised three collections. The first was a primary collection of 72 condition-specific clinical practice guidelines from the European Academy of Dermatology and Venereology (EADV). The second was a collection made from an authoritative dermatology reference work: Fitzpatrick's Dermatology in General Medicine (9th edition). The third was a collection of 7,936 dermatology case reports published between 2004 and 2024, which was included because guideline and textbook sources have sparse coverage of rare and atypical presentations. Raw text was extracted from PDF sources, stripped of reference lists and formatting artefacts, and chunks were embedded using Snowflake Arctic Embed V2 (Supplementary Table 1) and stored as dense vectors in ChromaDB.

### Retrieval pipeline

Retrieval proceeded as a bounded multi-step loop orchestrated by a Qwen3 model (Qwen3-235B-A22B-Instruct-2507; served through the Together API (https://api.together.ai/)). The set of collections available to the orchestrator was fixed per evaluation set (guideline and textbook for DermBenchQA and case reports for the DermCaseQA). Within that scope, for each question, the orchestrator first planned a retrieval strategy - which database collections to search, and whether to split the question into two to three focused sub-queries (prompts in Supplementary Notes 3.1 and 3.2). Then it retrieved the 50 highest-ranked candidate documents by dense similarity. It then issued a sufficiency judgement on the retrieved documents (prompt in Supplementary Note 3.3), returning a binary sufficient-or-insufficient decision and an explicit statement of the missing information if insufficient. Where context was judged insufficient, the orchestrator reformulated the query to target the identified knowledge gaps (prompt in Supplementary Note 3.4) and repeated retrieval, to a maximum of two additional iterations, giving three retrieval rounds in total. The loop terminated on a judgement of sufficiency or on reaching the iteration cap, whichever occurred first. Documents retrieved across iterations were pooled, deduplicated by chunk identifier, and reranked with Mixedbread Rerank Large v1 (mxbai-rerank-large-v1, hosted locally); the 10 highest-ranked passages were passed to the answer generation LLM as context.

### Generation and citation output

The generation model received the reranked passages prepended to the question prompt, each passage labelled with a numeric document identifier, and was instructed to answer using only the provided context and to list the identifiers of the passages used. On the multiple-choice set the model was asked to select the single best answer option and return only the corresponding letter; on the case-report-derived set, to give a single concise free-text answer. Prompts for both datasets, and for the corresponding no-retrieval baselines, are given in Supplementary Note 4.

## Benchmark evaluation

### Models evaluated

Five LLMs spanning a range of architectures, parameter scales and access modalities were evaluated: GPT-5 (OpenAI API), and the open-weight models Llama-4 Scout, Mistral Large 2, Qwen 2.5 and Gemma-3, hosted locally. Pipeline components: the embedding model, retrieval orchestrator, reranker and the annotating and adjudicating model, were fixed across all evaluated models. Full configurations are specified in Supplementary Table 5.

### Open-ended scoring

Open-ended responses were scored against the published reference diagnosis by Claude Sonnet 4.6 acting as an automated judge, using the prompt in Supplementary Note 5. The judge returned a verdict (correct, partially correct, incorrect), an answer counting as correct if it named the reference diagnosis or an accepted synonym at the specificity level of the reference. The judge was blinded to which model had produced each response.

## Reader study

### Preregistration and reporting

The reader-study protocol, hypotheses, outcomes, sampling procedure, and analysis plan were preregistered on the Open Science Framework before the start of data collection (https://osf.io/3e2ax/overview; registered 6 June 2026). The protocol was developed in accordance with SPIRIT-AI, and participant flow is reported in accordance with CONSORT-AI (Supplementary Fig. 1).

CORA's generation component used Llama-4 Scout, selected as the strongest open-weight performer on the multiple-choice benchmark from which reader-study items were drawn. "Open weights" was a prerequisite rather than a preference: on-premises inference and version pinning are necessary for a system handling identifiable patient data in routine care, and neither is available from API-only services. The reader study therefore evaluated a configuration a clinic could realistically deploy.

### Participants

Eligible participants were licensed physicians whose clinical work involved assessing dermatological presentations. No seniority threshold was applied beyond licensure. They were recruited by direct email invitation. Invitations described the study as an evaluation of an LLM decision-support tool and did not disclose that the model's answer or its citations were incorrect on a subset of items. Forty-six physicians completed all 16 cycles yielding 736 paired decisions.

### Question sampling

Questions for which the model cited more than five source passages were excluded from sampling, as reviewing an unusually large number of excerpts per question would have imposed a disproportionate reading burden on participants. These items constituted a small minority of the eligible pool (249 of 2,207 valid cases; 11.3%), and their exclusion did not introduce systematic bias in the direction that would matter: LLM accuracy on questions with more than five cited sources (89.6%) did not differ significantly from accuracy on the remaining questions (86.4%; Fisher's exact test, $P = 0.20$), and was in fact numerically higher, so exclusion could not have inflated the observed accuracy of the sampled cases.

### Study procedure

The study comprised three parts: a demographic section recording physician experience level; the study session, comprising 16 question cycles; and automatic submission. All items were mandatory and the platform prevented progression without a response, enforcing sequential completion. The estimated duration was 20-30 minutes.

Each cycle followed a fixed three-step sequence. First, the physician read the question stem and answer options and selected an unaided answer. Second, the physician was shown the same question and options together with CORA's selected answer and its cited source documents, and rated each cited document independently in response to the prompt: "Does this cited text contain enough information that supports the LLM's answer?", with response options "Supports" and "Does not support". Third, the physician decided whether to keep or revise the original answer.

Support ratings were therefore collected after the model's answer was revealed. Physicians retained full autonomy over the final answer; the model output was advisory, no feedback mechanism existed, and physicians were never told whether a recommendation was correct. Participants were instructed not to consult colleagues, textbooks or other external resources during the session; this was a behavioural instruction that could not be technically verified given the remote design.

Cited documents were presented with topically relevant sentences highlighted, to prevent grounding ratings from being recorded as unsupported merely because relevant content was overlooked within a long document, and to bound reading burden across 16 cycles. Highlighting was generated by the same Llama-4 Scout model using a pre-specified prompt (Supplementary Note 6) excluding sentences irrelevant to the question stem and to all answer options, blinded to the answer option the model had selected, so that highlighting could not directionally favour the model's answer. Source metadata, including guideline name and textbook title, were withheld, so that ratings would reflect the textual relationship between document and answer rather than institutional trust in the originating source.

There was no conventional-resources comparator arm: all participants completed the same unaided-then-assisted sequence, and no arm received guidelines or search tools in the absence of CORA. The accuracy change reported here therefore bounds the benefit attributable to CORA specifically rather than isolating it from the benefit of decision support in general.

## Statistical analysis

Analyses were performed in Python 3.12.3. Tests were two-sided at $\alpha=0.05$ unless stated otherwise, and 95% confidence intervals were obtained by bootstrap resampling (10,000 replicates, seed 42). P values across related hypotheses were Holm–Bonferroni corrected. For each backbone model, CORA was compared against its no-retrieval baseline using a one-sided Wald non-inferiority test (margin 1 pp) with item-level bootstrap CIs. Per-physician unaided versus assisted accuracy was compared with a Wilcoxon signed-rank test. Associations between citation support and correctness were expressed as odds ratios, estimated at the decision level by logistic regression with standard errors clustered by physician, and at the question level for CORA's own correctness. Relative AI reliance and relative self-reliance, overall and stratified by citation support, are reported with physician-clustered bootstrap CIs. Reader-study CIs throughout account for within-physician dependence by resampling physicians with replacement.

## Statement on the use of AI tools

During manuscript preparation, the authors used AI tools to improve the clarity, wording, and organization of author-written text. All outputs were critically reviewed and revised by the authors, who take full responsibility for the final manuscript.

## Data availability statement

The benchmark and case reports question sets and the reader study data are available on figshare at https://figshare.com/s/fa2ff8eb45984acfecf6. The source question corpora and case reports are publicly available as described in the Methods.

## Code availability statement

The source code used to implement CORA and conduct the analyses reported in this study is publicly available at https://github.com/DBO-DKFZ/CORA. The repository contains the code for document ingestion and indexing, agentic retrieval, answer generation, evaluation and statistical analysis, together with the associated configuration files and computational notebooks.

## Consortium

Alexander Salava, Alexander Thiem, Ana Sanader Vučemilović, Andrea Miyuki Yoshimura, Andzelka Ilieva, Arzu Ferhatosmanoğlu, Aude Beyens, Claudia Pföhler, Dilara Ilhan Erdil, Florentia Dimitriou, Isabelle Hoorens, Janik Fleißner, Johan Dahlberg, Juan Sebastián Andreani Figueroa , Julia Welzel, Katerina Damevska, Kristine, Elisabeth Mayer, Lara Valeska Maul, Laura Garzona-Navas, Laura Isabell Bley, Lidija Petrovska, Martha, Alejandra Morales-Sánchez, Martyna Sławińska, Miroslav Dragolov, Nina Booken, Nkechi Anne Enechukwu, Oana-Diana Persa, Panagiota Theofilogiannakou, Roque Rafael Oliveira Neto, Rym Afiouni, Sandra Schuh, Sonia Rodriguez Saa , Stefana Damevska, Vanda Bondare-Ansberga, Verena Ahlgrimm-Siess, Viktor Simeonovski, Zsuzsanna Lengyel, Sandra Peternel, Emmanouil Chousakos, Wiebke K. Peitsch, İrem Özdemir, Markus Thieme, Alexandra Chrusciak

## Conflicts of Interests

J.N.K. declares consulting services for Bioptimus, France; Panakeia, UK; AstraZeneca, UK; and MultiplexDx, Slovakia. Furthermore, he holds shares in StratifAI, Germany, Synagen, Germany, Ignition Lab, Germany; has received an institutional research grant by GSK; and has received honoraria by AstraZeneca, Bayer, Daiichi Sankyo, Eisai, Janssen, Merck, MSD, BMS, Roche, Pfizer and Fresenius. T.J.B. is the owner of Smart Health Heidelberg GmbH, a software company which develops digital health apps and received honoraria from Novartis, Roche, Merck and HEINE Optotechnik. The other authors declare no competing interests.

## Author Contributions

TC conceived of and designed the overall study. TJB, TC were responsible for reader study participant recruitment. TC developed the system with dermatological expert feedback from TJB, CNG. TC compiled the data with assistance from MJH. TC set up and conducted the web-based reader study, and conducted experimental evaluation. TC conducted statistical data analysis with expert feedback from CW. TC generated the figures. FS, NBM and all other authors provided clinical and/or machine learning expertise and contributed to the interpretation of the results. TC wrote the manuscript with input from all authors. TJB led the conceptualization, provided resources, reviewed and edited the manuscript. All authors reviewed and corrected the final manuscript and collectively made the decision to submit for publication.

# Extended Data

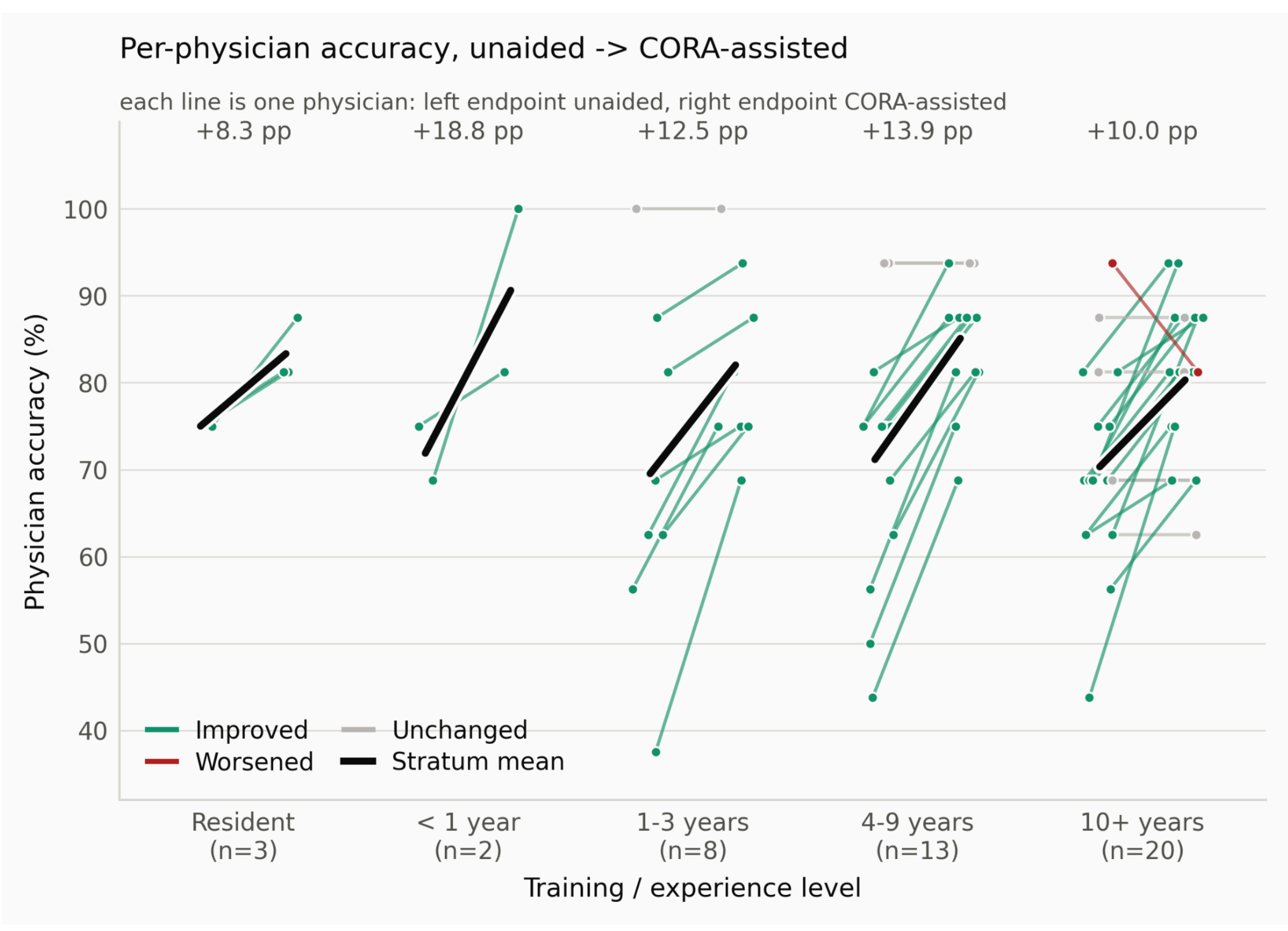

**Extended Data Figure 1: Accuracy by experience levels**

Per-physician accuracy gains with LLM assistance by experience level. Each line connects one physician's unaided (left) and CORA-assisted (right). Mean gains (pp) are shown above each stratum. Improvement was consistent at every level of training and experience (n = 46).

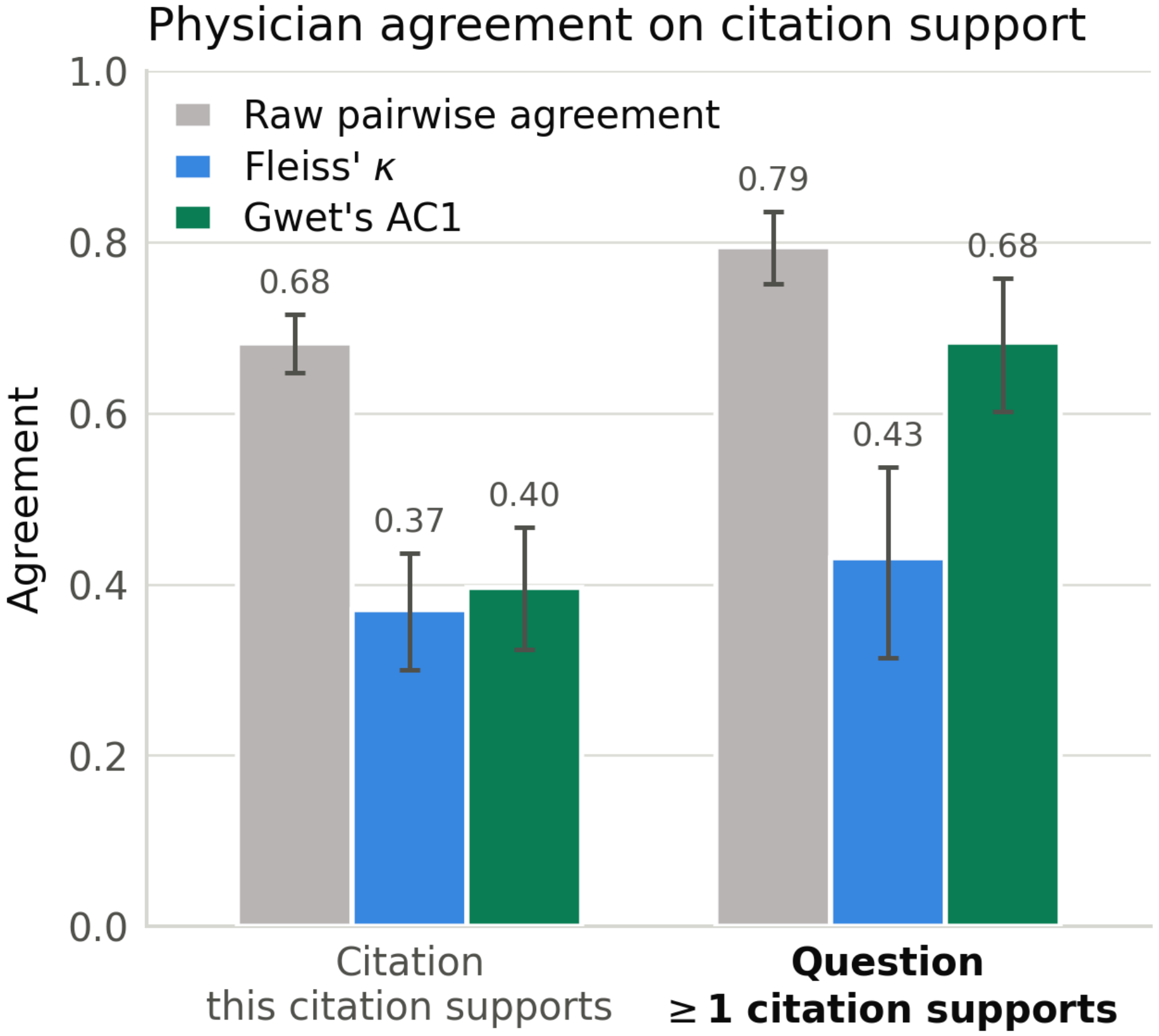

**Extended Data Figure 2: Inter-rater agreement**

Agreement per citation is low-moderate but agreement per question is moderate-high.